\documentclass[letterpaper, 10 pt, conference]{ieeeconf}  % Comment this line out if you need a4paper

\IEEEoverridecommandlockouts                              % This command is only needed if 
\usepackage{amsmath} % assumes amsmath package installed
\usepackage{amssymb}  % assumes amsmath package installed
\usepackage{bm}
\usepackage{hyperref}
\usepackage{graphicx}

\newcommand{\symbf}[1]{\bm{#1}}

\title{\LARGE \bf
Generating Synthetic Ground Truth Distributions for Multi-step Trajectory Prediction using Probabilistic Composite Bézier Curves
}

\author{Ronny Hug$^{\dagger}$ and Stefan Becker$^{\dagger}$ and Wolfgang H\"ubner$^{\dagger}$ and Michael Arens$^{\dagger}$% <-this % stops a space
\thanks{$^{\dagger}$Fraunhofer IOSB, Fraunhofer Center for Machine Learning, Ettlingen, Germany.
        {\tt\small firstname.lastname@iosb.fraunhofer.de}}%
}

\begin{document}

\maketitle
\thispagestyle{empty}
\pagestyle{empty}

%%%%%%%%%%%%%%%%%%%%%%%%%%%%%%%%%%%%%%%%%%%%%%%%%%%%%%%%%%%%%%%%%%%%%%%%%%%%%%%%
\begin{abstract}
     
        An appropriate data basis grants one of the most important aspects for training and evaluating probabilistic trajectory prediction models based on neural networks.
        In this regard, a common shortcoming of current benchmark datasets is their limitation to sets of sample trajectories and a lack of actual ground truth distributions, which prevents the use of more expressive error metrics, such as the Wasserstein distance for model evaluation.
        Towards this end, this paper proposes a novel approach to synthetic dataset generation based on composite probabilistic Bézier curves, which is capable of generating ground truth data in terms of probability distributions over full trajectories.
        This allows the calculation of arbitrary posterior distributions. 
        The paper showcases an exemplary trajectory prediction model evaluation using generated ground truth distribution data.

\end{abstract}

%%%%%%%%%%%%%%%%%%%%%%%%%%%%%%%%%%%%%%%%%%%%%%%%%%%%%%%%%%%%%%%%%%%%%%%%%%%%%%%%

\section{Introduction}
An integral component for training and evaluating neural network models is the use of an appropriate data basis.
Taking multi-step human trajectory\footnote{Here, a trajectory is defined as a sequence of locations along a path with some velocity profile attached to it.} prediction as an example, where given a sequence of spatial positions $\{\mathbf{x}_1, ... \mathbf{x}_N\}$, the observation, the next $M$ positions $\{\mathbf{x}_{N+1}, ..., \mathbf{x}_{N+M}\}$ are to be predicted, obtaining such a data basis is especially difficult.
This is because an adquate representation for the future progression of an observed trajectory is given by a conditional multi-modal probability distribution $p(\mathbf{x}_{N+1}, ..., \mathbf{x}_{N+M}|\mathbf{x}_1, ... \mathbf{x}_N)$, hence favors the use of probabilistic prediction models, which learn to approximate this distribution.
In this case an ideal data basis would provide these conditional ground truth distributions.
However, there is no reliable way of deriving required conditional distributions from commonly used trajectory datasets, e.g. provided by benchmarks such as \emph{Thör} \cite{rudenko2020thor} or \emph{TrajNet++} \cite{kothari2020trajnetpp}, only leaving the option of resorting to synthetically generated data in case distribution-based ground truth data is desired\footnote{E.g. in case an evaluation aims at a more nuanced model evaluation not achievable by the common approach of using the negative log-likelihood or top-k metrics together with plain trajectory data.}.

A common approach for synthetically generating trajectory data consists of first generating paths through a virtual scene, either using waypoints or simple motion patterns.
Virtual agents then follow these paths, optionally complying with physical constrains (e.g. \cite{becker2022uav}) or interacting with other agents \cite{helbing1995social}, whereby their trajectory is recorded.
Finally, sensor or annotation noise is simulated by applying a noise model to each individual trajectory point.
%In this way, data can be obtained in a controlled way with specific constrains given by design.
However, this approach of injecting uncertainty often limits the dataset to representing probability distributions on a per point basis instead of distributions over full trajectories, due to a lack of knowledge about intra-trajectory and inter-path correlations.  

Towards this end, this paper proposes a novel approach to synthetic trajectory data generation for probabilistic trajectory prediction models, which is capable of generating ground truth data in terms of probability distributions over full trajectories. The approach utilizes probabilistic (composite) Bézier curves ($\mathcal{N}$-Curves, \cite{hug2020introducing}) for modeling individual paths and arranges multiple curves in a mixture distribution for building multi-path datasets.
By exploiting the $\mathcal{N}$-Curve's equivalence with Gaussian processes, datasets defined this way enable the calculation of conditional distributions over trajectories given arbitrary observations and thus the use of more expressive performance metrics, such as the Wasserstein distance alongside the commonly used negative log-likelihood. 
In order to showcase the application of the proposed approach for benchmarking probabilistic trajectory prediction models, an exemplary evaluation following the common benchmarking approach is provided.

\section{Dataset Generation Approach}
In this paper, the primary goal is to derive an approach for trajectory dataset generation, which yields a probability distribution over full trajectories covering multiple paths through a virtual (structured) environment and further allows for the calculation of conditional distributions during training or test given different observations. 
The proposed dataset generation approach consists of multiple stages, which are explained in more detail in the following:
\begin{enumerate}
        \item Definition of paths through a virtual (structured) environment in terms of probabilistic Bézier curves.
        \item Definition of velocity profiles by curve discretization.
        \item Derivation of the dataset prior distribution.
        \item As required: Calculation of posterior distributions given specific observed trajectories.
\end{enumerate}
A variant of this dataset generator is implemented and utilized in the scope of the latest version\footnote{Building on the core concepts presented in the initial STSC paper \cite{hug2020complementary}} of the STSC benchmark and is available at \url{https://github.com/stsc-benchmark/stsc-lib}.  

\paragraph{Defining paths through a virtual (structured) environment}
In order to provide an intuitive way of defining paths within a dataset, $\mathcal{N}$-Curves \cite{hug2020introducing}, a probabilistic extension of Bézier curves \cite{prautzsch2002bezier} which add uncertainty to points along the curve, are chosen as the basic building block for representing individual paths. 
%These provide a compact representation for spatial curves with uncertainty, have an intuitive construction approach and are numerically stable.
By stringing together multiple $\mathcal{N}$-Curves into a composite curve, complex paths can easily be pieced together.

\begin{figure}[h]
        \includegraphics[width=0.95\columnwidth]{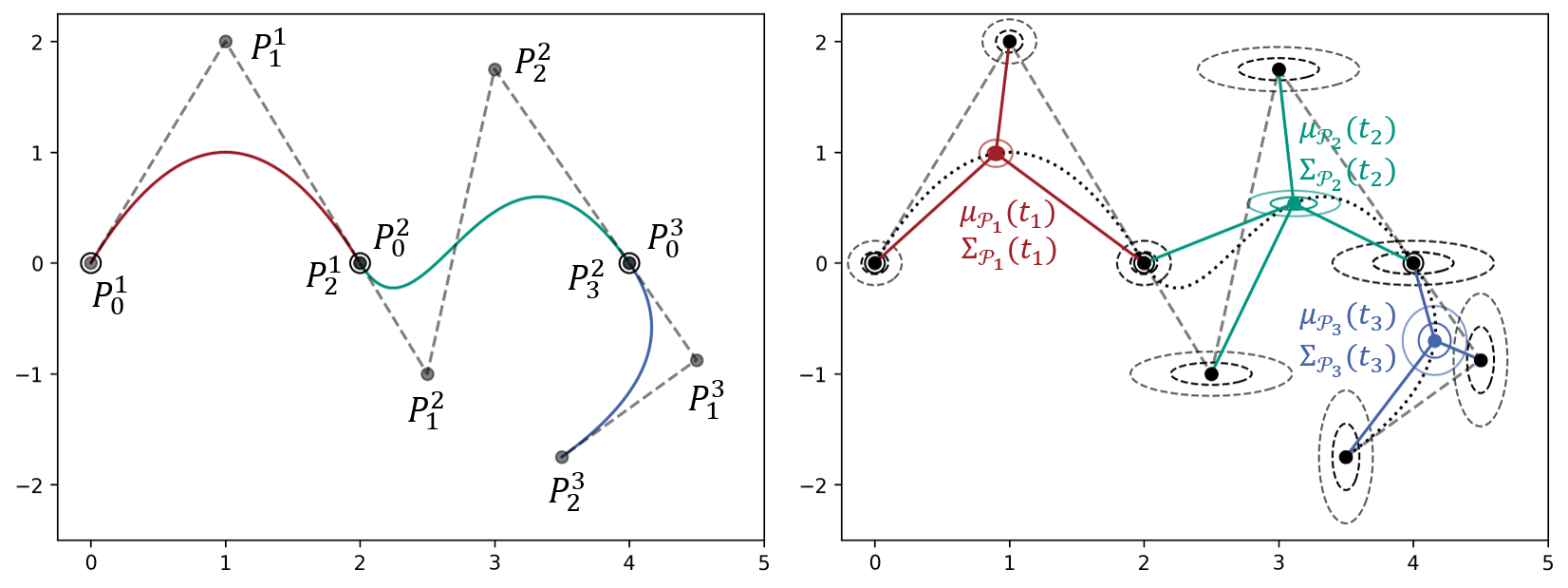}
        \caption{Exemplary composite $\mathcal{N}$-Curve consisting of $N_{seg} = 3$ segments with control point sets $\mathcal{P} = \{\mathcal{P}_1, \mathcal{P}_2, \mathcal{P}_3\}$, where $\mathcal{P}_1 = \{P^1_0, P^1_1, P^1_2\}$, $\mathcal{P}_2 = \{P^2_0, P^2_1, P^2_2, P^2_3\}$ and $\mathcal{P}_3 = \{P^3_0, P^3_1, P^3_2\}$, with $L_1 = L_3 = 2$ and $L_2 = 3$. Each control point $P^j_l \sim \mathcal{N}(\cdot|\cdot)$ follows a Gaussian distribution with respective mean $\mu^j_l$ and covariance matrix $\Sigma^j_l$. Left: The resulting mean curve with control point locations. Covariance ellipses are omitted for clarity. Right: Gaussian curve points along the composite $\mathcal{N}$-Curve at curve positions $t_1 = 0.15$, $t_2 = 0.55$ and $t_3 = 0.8$. The influence of control points on each curve point is indicated by solid lines.} %Each segment occupies an equal range in $[0, 1]$, i.e. the $j$'th segment covers $[\frac{j-1}{N_{seg}},\frac{j}{N_{seg}}]$.}
        \label{fig:prob_spline}
\end{figure}

Formally, $\mathcal{N}$-Curves, defined by $(L + 1)$ independent $d$-dimensional Gaussian control points $\mathcal{P} = \{P_0, ..., P_L\}$ with $P_l \sim \mathcal{N}(\symbf{\mu}_l,\symbf{\Sigma}_l)$, are employed as a foundation for a pattern-based trajectory dataset description.
Through the curve construction function
\begin{align}
	\label{eq:BN}
	X_t = B_{\mathcal{N}}(t, \mathcal{P}) = (\mu_\mathcal{P}(t), \Sigma_\mathcal{P}(t)),~\text{with}
\end{align}
\begin{align}
        \label{eq:mu_psi}
        \mu_\mathcal{P}(t) = \sum_{l=0}^{L} b_{l,L}(t) \symbf{\mu}_l~\text{and}~\Sigma_\mathcal{P}(t) = \sum_{l=0}^{L} \left(b_{l,L}(t)\right)^2 \symbf{\Sigma}_l,
\end{align}
where $b_{l,L}(t) = \binom{L}{l} (1-t)^{L-l}t^l$ are the Bernstein polynomials \cite{lorentz2013bernstein}, the stochasticity is passed from the control points to the curve points $X_t \sim \mathcal{N}(\mu_\mathcal{P}(t),\Sigma_\mathcal{P}(t))$, yielding a sequence of Gaussian distributions $\{X_t\}_{t \in [0,1]}$ along the underlying B\'ezier curve.
Here, the curve parameter $t \in T=[0,1]$ indicates the position on the curve.
Stringing together $N_{seg}$ $\mathcal{N}$-Curves with respective Gaussian control points $\mathcal{P}_j = \{P^j_0, ..., P^j_{L_j}\}$, where $j \in \{1,...,N_{seg}\}$, yields a composite curve.
In this case, $t$ still remains in $[0, 1]$ and traverses all segments of the composite curve.
For curve point calculation, only the control points of the segment the curve point resides in are used. 
The segment is determined from $t$ by the mapping $j = m_c(t) = \max\{1, \lceil \frac{t}{\tau} \rceil\}$ with $\tau = \frac{1}{N}$. %, which maps $t$ onto the corresponding segment $j$.
For segment-specific calculations $t$ is mapped onto the segment local position $\frac{t - m_c(t) \cdot \tau}{(m_c(t)+1) \cdot \tau - m_c(t) \cdot \tau} = \text{loc}(t) \in [0, 1]$. 
%Further, when defining $\mathcal{N}$-Curve segments, the connecting point is required to have equal mean vectors and covariance matrices.
%Lastly, continuity constrains can be implemented in the curve by introducing geometric dependencies on control points close to the connecting point of subsequent segments.
%This aspect is discussed later. 
Fig. \ref{fig:prob_spline} depicts a composite $\mathcal{N}$-Curve.

Finally, for modeling multiple paths in a single dataset, $K$ composite $\mathcal{N}$-Curves with respective control point sets $\mathcal{P}_k$ can be combined into a mixture with prior weight distribution $\pi = \{\pi_1, ... \pi_K\}$ over the curves.
In this way, complex datasets can be represented, where each mixture component models a specific path through the scene.

\paragraph{Defining velocity profiles by $\mathcal{N}$-Curve mixture discretization}
In order to extract trajectory data from a set of paths defined in terms of a mixture of $K$ (continuous) composite $\mathcal{N}$-Curves, discrete subsets $T^k_N$ of $T$ can be employed for extracting length $N$ trajectories with (Gaussian) trajectory points $X_i = B_\mathcal{N}(t_i, \mathcal{P}_k)~\forall t_i \in T^k_N$ from single mixture components.
Altering the relative placement of the $t_i$ yields different trajectory velocity profiles, e.g. reflecting constant or accelerating movement speed. 
For example, constant speed can be achieved by setting $T^k_N$, such that constant distance $\|\mu_{\mathcal{P}_k}(t_{i+2}) - \mu_{\mathcal{P}_k}(t_{i+1})\| = \|\mu_{\mathcal{P}_k}(t_{i+1}) - \mu_{\mathcal{P}_k}(t_{i})\|$ between subsequent curve points is achieved.
The curve parameter subsets can be defined on a per path basis and be of varying length. 

\paragraph{Deriving the dataset's prior distribution}
Given a dataset defined in terms of a mixture of composite $\mathcal{N}$-Curves and a discrete curve parameter subset $T^k_{N_k}$\footnote{$T^k_{N_k}$ denotes the finite curve parameter subset of the $k$'th mixture component, covering trajectories of length $N_k$.} for each mixture component, the dataset's prior distribution over the set of possible trajectories has to be calculated.
This can be achieved by exploiting the equivalence of $\mathcal{N}$-Curves and a specific class of Gaussian processes (GP) \cite{hug2022ngp} in order to derive a vector-valued mean and matrix-valued covariance function from each mixture component.
This ultimately converts the mixture of composite $\mathcal{N}$-Curves into a mixture of GPs.
Using this mixture of GPs, a prior distribution modeling trajectories with lengths $N_k$ can be derived in terms of a Gaussian mixture distribution with weights $\{\pi_k\}_{k \in \{1,..,K\}}$, mean vectors $\{\mu_k\}_{k \in \{1,..,K\}}$ and covariance matrices $\{\Sigma_k\}_{k \in \{1,..,K\}}$. 
The derivation of the $\mu_k$ and $\Sigma_k$ is detailed in the following.
%For keeping this paragraph as concise as possible, the derivation focuses on a single mixture component and thus the component index $k$ is omitted in the following in order to reduce visual clutter.
For conciseness, the derivation focuses on a single mixture component and omits the component index $k$ for reducing visual clutter.

Starting with the vector-valued mean function $\mathbf{m}_{\mathcal{P}}$, each mixture component's prior mean vector $\mu$ is given by the concatenation of all mean vectors $\symbf{\mu}_{\mathcal{P}}(t_i)$ (see Eq. \ref{eq:mu_psi}) along the underlying Bézier curve, i.e.
\begin{align}
	\mu = (\mathbf{m}_{\mathcal{P}}(T_N))^\top = 
	\begin{pmatrix}
		(\symbf{\mu}_{\mathcal{P}}(t_1))^\top &
		\cdots &
		(\symbf{\mu}_{\mathcal{P}}(t_N))^\top
	\end{pmatrix}.
\end{align}
In the case of 2d trajectories, the resulting vector consists of $N_k$ 2d vectors, yielding a $(2 \cdot N_k \times 1)$ vector.

The covariance matrix $\Sigma$ is given by the block-partitioned Gram matrix calculated from the respective matrix-valued covariance function $\mathbf{K}_{\mathcal{P}}(t_{i_1},t_{i_2})$ with $t_{i_1},t_{i_2} \in T_{N}$, which connects\footnote{In terms of their correlation.} two Gaussian curve points $X$ and $Y$ on the same composite $\mathcal{N}$-Curve, where $X = B_\mathcal{N}(t_{i_1},\mathcal{P}_{m_c(t_{i_1})})$ and $Y = B_\mathcal{N}(t_{i_2},\mathcal{P}_{m_c(t_{i_2})})$.
With $X$ and $Y$ potentially residing on different curve segments, and thus potentially being calculated from different control points, multiple cases have to be considered when deriving $\mathbf{K}_\mathcal{P}(t_{i_1},t_{i_2})$\footnote{Here, $t_{i_1} \leq t_{i_2}$ is assumed.}:
\begin{enumerate}
        \item $m_c(t_{i_1}) = m_c(t_{i_2})$: $X$ and $Y$ reside on the same $\mathcal{N}$-Curve segment. This is the case covered by \cite{hug2022ngp}.
        \item $m_c(t_{i_1}) + 1 = m_c(t_{i_2})$: $X$ and $Y$ reside on subsequent $\mathcal{N}$-Curve segments.
        \item $m_c(t_{i_1}) + 1 < m_c(t_{i_2})$ There is at least one curve segment in between the segments both. 
\end{enumerate}

\begin{figure}[h]
        \includegraphics[width=0.95\columnwidth]{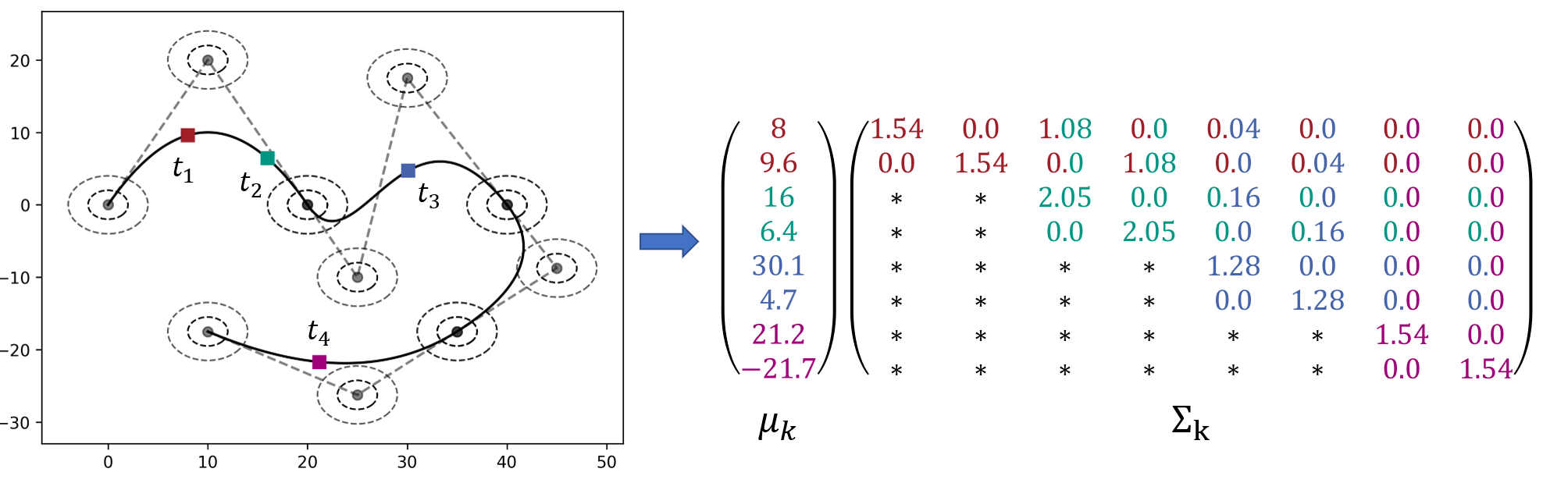}
        \caption{Example of a mean vector and covariance matrix derived from an $\mathcal{N}$-Curve covering $N=4$ Gaussian curve points.}
        \label{fig:prior}
\end{figure}

In case both $X$ and $Y$ reside on the same segment (see e.g. $t_1$ and $t_2$ in Fig. \ref{fig:prior}), the covariance function
\begin{align}
\begin{split}
\label{eq:kernel}
        \mathbf{K}_{\mathcal{P}_{m_c(t_{i_1})}}&(t_{i_1},t_{i_2}) = \mathbb{E}[(\bm{X} - \mu_X)(\bm{Y} - \mu_Y)^\top] \\
        &= \mathbb{E}\left[ \bm{X}\mathbf{Y}^\top \right] - \mu_X\mu^\top_Y\\
        &= \sum^{L}_{l=0} b_{l,L}(\text{loc}(t_i)) b_{l,L}(\text{loc}(t_j)) \left( \Sigma_l + \mu_l \mu^\top_l \right) \\
        &+ \sum^{L}_{l=0} \left( \sum^{L}_{l'=0,l' \neq l} b_{l,L}(\text{loc}(t_i)) b_{l',L}(\text{loc}(t_j)) \mu_l \mu^\top_{l'} \right) \\
        &-\mu_X\mu^\top_Y,
\end{split}
\end{align}
derived in \cite{hug2022ngp} can be used. % without modification.
Here, $\mu_X$ and $\mu_Y$ denote the mean vectors of $X$ and $Y$.
The calculation is based on the $m_c(t_{i_1})$'th segment's Gaussian control points $\mathcal{P}_{m_c(t_{i_1})}$.

\begin{figure}[h]
        \includegraphics[width=0.95\columnwidth]{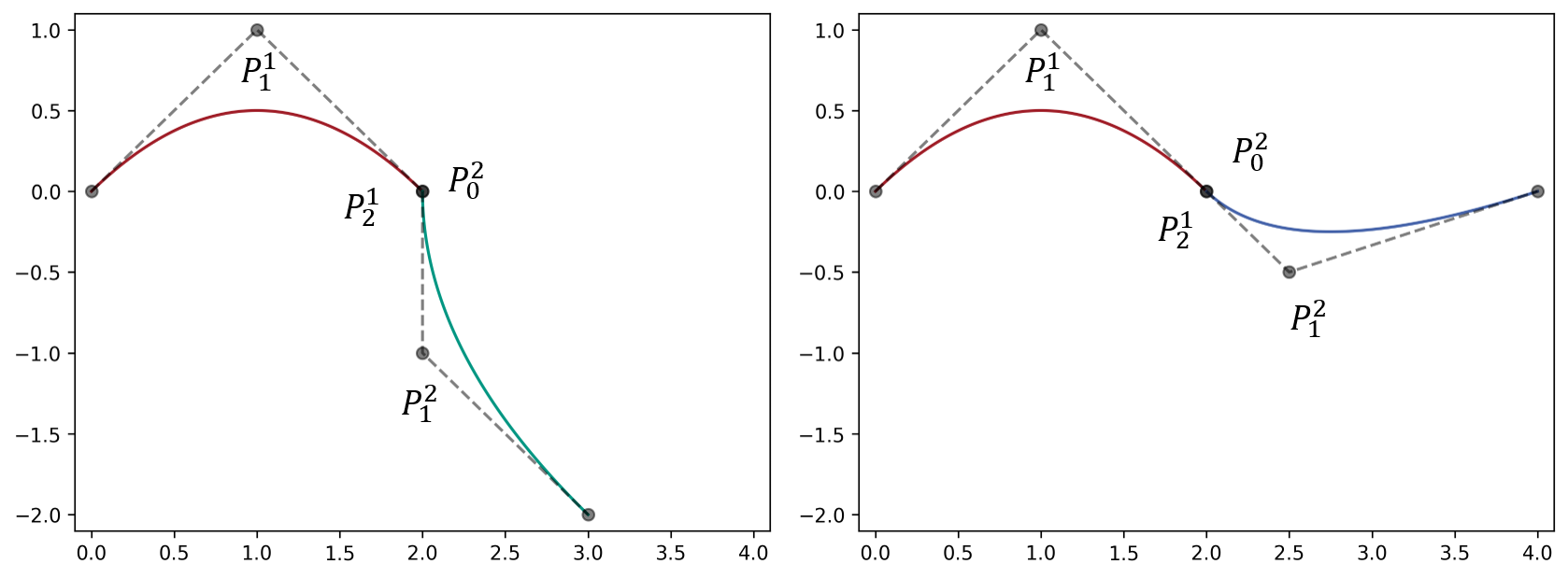}
        \caption{Example for a $C^0$ (left) and a $C^1$ (right) continuous curve.}
        \label{fig:continuity}
\end{figure}

In case $X$ and $Y$ reside on subsequent segments (see e.g. $t_2$ and $t_3$ in Fig. \ref{fig:prior}), Eq. \ref{eq:kernel} has to be modified to consider control point dependencies emerging from the composite curve's continuity in its connecting points.
While $C^0$ continuity is inherent, $C^1$ and $C^2$ continuity emerge from equal tangency ($C^1$ and $C^2$) together with equal curvature ($C^2$) in the connecting point for two connected segments.
Geometrically, considering $2$ connected segments $j$ and $j+1$, $C^1$ continuity can be enforced by arranging $P^j_{L_j-1}$, $P^j_{L_j} = P^{j+1}_0$ (connecting point) and $P^{j+1}_1$ on a straight line.
%Geometrically, $C^1$ continuity can be enforced by arranging the second-last control point of a segment $P^j_{L_j-1}$, the connecting point $P^j_{L_j} = P^{j+1}_0$ and the second control point of the following segment $P^{j+1}_1$ on a straight line.
$C^2$ continuity is given for equidistant points, i.e. $\|P^j_{L_j} - P^j_{L_j-1}\| = \|P^{j+1}_1 - P^j_{L_j}\|$.
An example for a $C^0$ and a $C^1$ continuous composite curves is depicted in Fig. \ref{fig:continuity}.
Following this, given $C^1$ or $C^2$ continuity, correlations between curve points of subsequent segments emerge through the geometric dependency of the control point sets, given by $P^{j+1}_1 = P^j_{L_j} + s \cdot (P^j_{L_j} - P^j_{L_j-1})$ with $s = \frac{\|P^{j+1}_{1} - P^{j+1}_{0}\|}{\|P^j_{L_j} - P^j_{L_j-1}\|}$.
%P_j[1] = P_i[-1] + s * (P_i[-1] - P_i[-2])
%s = np.linalg.norm(j_off) / np.linalg.norm(i_off), i_off = P_i[-1] - P_i[-2], j_off = P_j[1] - P_j[0]
Hence the calculation of $\mathbb{E}[\bm{X}\bm{Y}^\top]$, which combines all Gaussian control points of both segments $m_c(t_{i_1})$ and $m_c(t_{i_2})$, is adjusted accordingly:
\begin{align}
\begin{split}
        \mathbb{E}[\bm{X}\bm{Y}^\top] &= \sum_{l_1=0}^{L_1} \sum_{l_2=0}^{L_2} b_* \\
        &\cdot \begin{cases}
                (\Sigma_{l_1} + \mu_{l_1}\mu_{l_2}^\top), & c_1 \\
                (\mu_{l_1}\mu_{L_1}^\top + s \cdot (\mu_{l_1}\mu_{L_1}^\top - \mu_{l_1}\mu_{l_1}^\top)), & c_2 \\
                (\mu_{L_1}\mu_{L_1}^\top + s \cdot (\mu_{L_1}\mu_{L_1}^\top - \mu_{L_1}\mu_{L_1-1}^\top)), & c_3 \\
                \mu_{l_1}\mu_{l_2}^\top, & c_4
        \end{cases}.
\end{split}
\end{align}
Here, $b_* = b_{l_1,L_1}(\text{loc}(t_{i_1})) b_{l_2,L_2}(\text{loc}(t_{i_2}))$ is the weighting factor and the cases $c_1$ through $c_4$ are given by
\begin{align*}
\begin{split}
        c_1 &\equiv l_1=L_1 \land l_2=0~\text{(connecting point)} \\ 
        c_2 &\equiv l_1=L_1-1 \land l_2=1 \\
        c_3 &\equiv l_1=L_1 \land l_2=1 \\
        c_4 &\equiv \text{else (independent points)}.
\end{split}
\end{align*}
The outer ($\sum_{l_1=0}^{L_1}$) and inner ($\sum_{l_2=0}^{L_2}$) sums  iterate over the control points of segment $m_c(t_{i_1})$ and $m_c(t_{i_2})$, respectively.

%Lastly, in case $X$ and $Y$ do not reside on the same or directly connected segments,
Lastly, in case $X$ and $Y$ reside on disconnected segments,
\begin{align} 
        \mathbb{E}[\bm{X}\bm{Y}^\top] = \sum_{l_1=0}^{L_1} \sum_{l_2=0}^{L_2} b_{l_1,L_1}(\text{loc}(t_{i_1})) b_{l_2,L_2}(\text{loc}(t_{i_2})) \mu_{l_1}\mu_{l_2}^\top
\end{align} 
collapses into the independent points case ($c_4$).
A schematic illustrating the three cases with the resulting prior mean vector and covariance matrix is depicted in Fig. \ref{fig:prior}.

\begin{figure}[h]
        \centering
        \includegraphics[width=0.48\columnwidth]{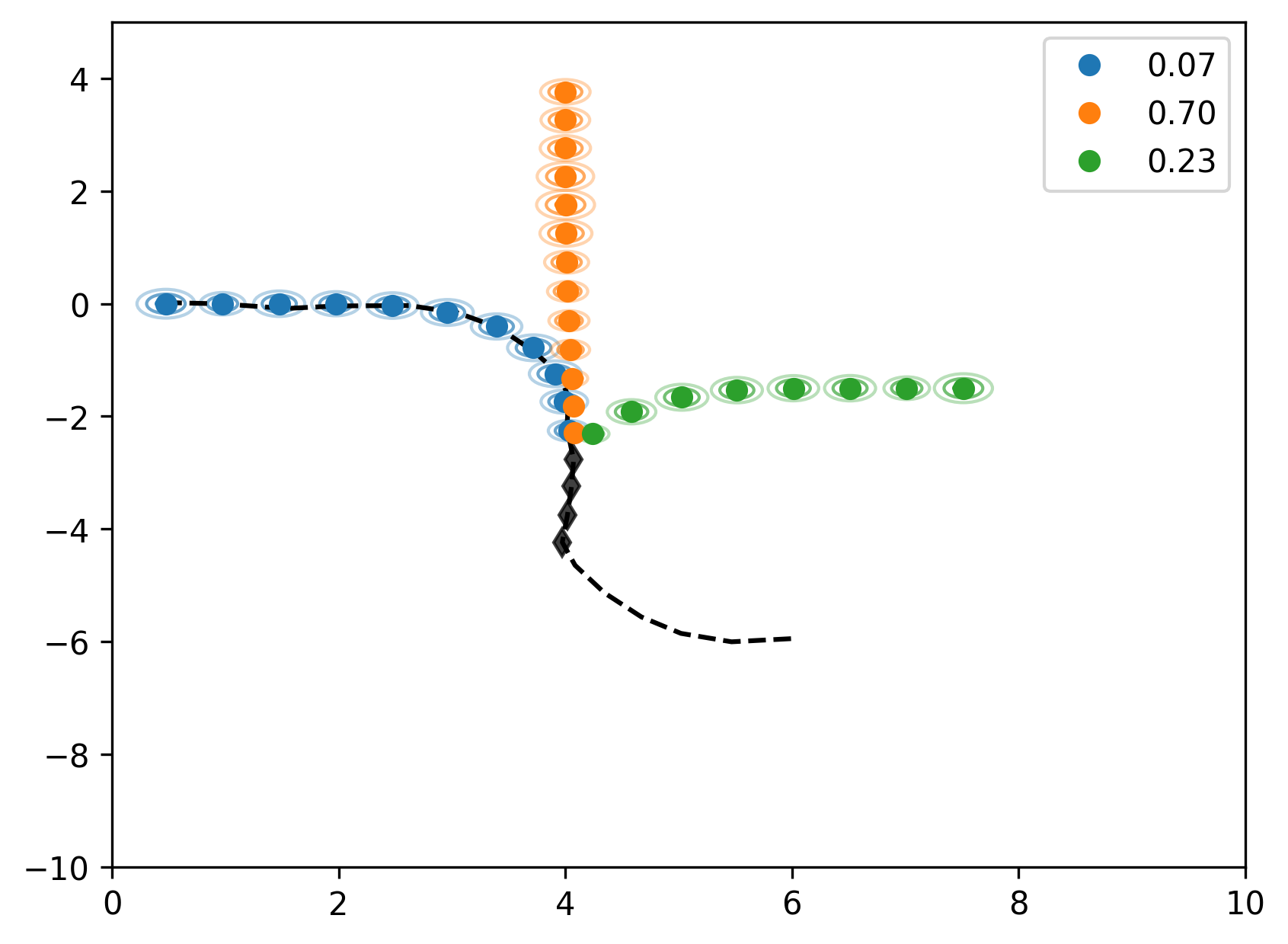}
        \includegraphics[width=0.48\columnwidth]{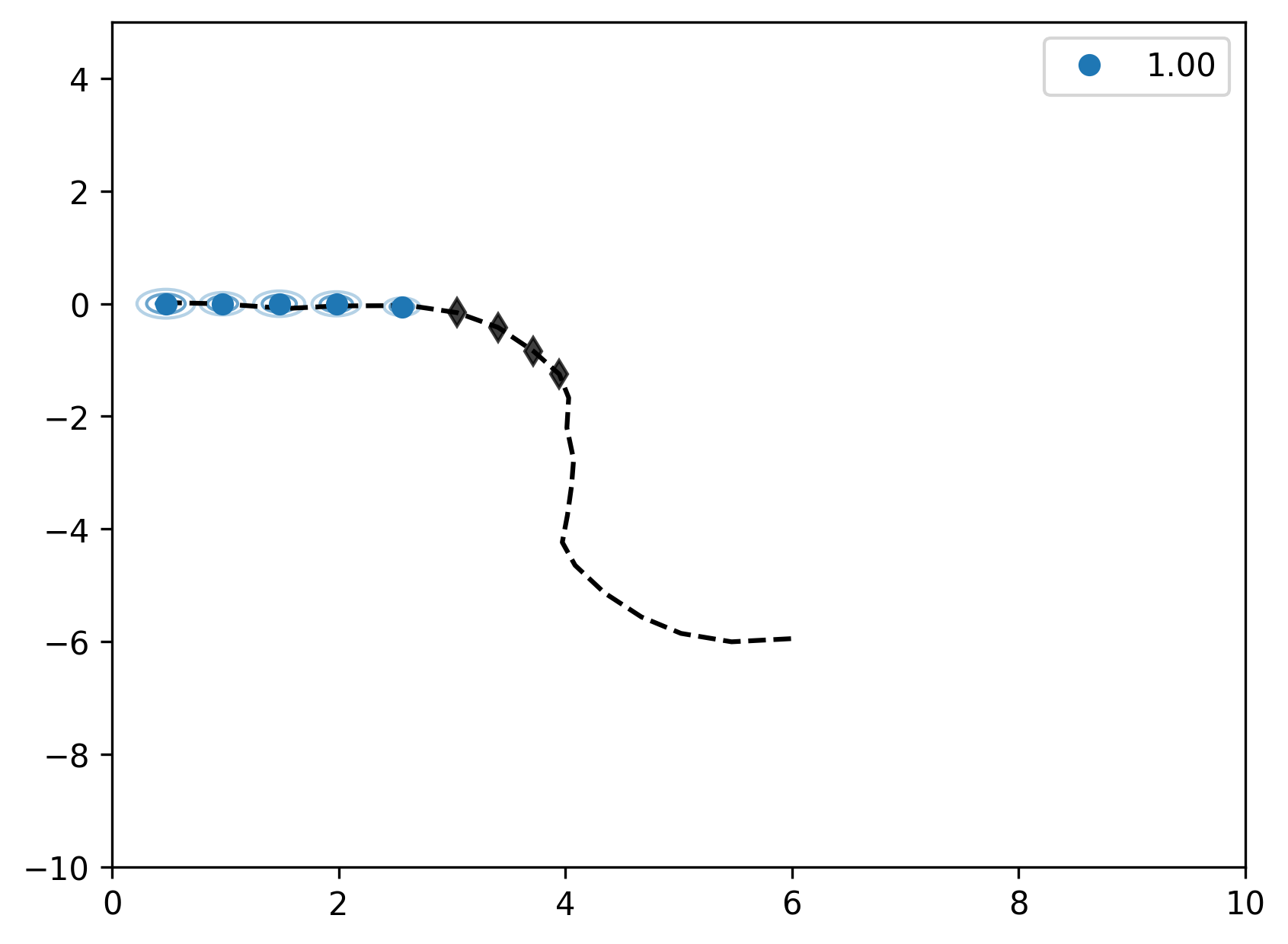}
        \caption{Posterior distributions given by conditioning on different subsets of the same trajectory. Fig. \ref{fig:dataset} depicts the underlying prior distribution. }
        \label{fig:posterior_examples}
\end{figure}

\paragraph{Calculating posterior distributions}
While the prior distribution suffices for generating trajectory data through sampling, calculating posterior distributions given different observations is the final missing piece for obtaining conditional ground truth distributions.
The Gaussian mixture-based prior distribution models full trajectories of length $N$\footnote{For simplicity equal trajectory length is assumed for each component.}, which yields a joint probability distribution $p(X_1, ..., X_N)$ over the trajectory points.
By partitioning the joint prior distribution into a partition containing the $N_\text{obs}$ observed time steps to condition on and the remaining $N_\text{pred}$ time steps, i.e. $p(X_1,...,X_N) = p(\{X_1,...,X_{N_\text{obs}}\} \cup \{X_{N_\text{in}+1},...,X_N\}) = p(X_A \cup X_B)$, the conditional distribution $p(X_B|X_A)$ can be calculated directly (see e.g. \cite{petersen2008matrix,pml1Book}).
The probability distribution for individual trajectory points can be extracted through marginalization
An example for different posterior distributions using different subsets of the same sample trajectory is depicted in Fig. \ref{fig:posterior_examples}.

\section{Exemplary Evaluation based on Gaussian Mixture Datasets and the Wasserstein Distance}
This section provides a brief showcase on how the Gaussian mixture-based dataset can be used within the standard evaluation approach for employing the Wasserstein distance alongside the commonly used negative log-likelihood (\emph{NLL}, \cite{ivanovic2019trajectron}). 
The Wasserstein distance \cite{kolouri2017optimal} $W_p(P,Q)$ quantifies the dissimilarity between two probability distributions $P$ and $Q$ by measuring the work required to transport the probability mass from $P$ to $Q$.
For dimensions $d > 1$, an approximation, such as the sliced Wasserstein distance \cite{bonneel2015sliced}, has to be used.

\begin{figure}[h]
        \includegraphics[width=0.48\columnwidth]{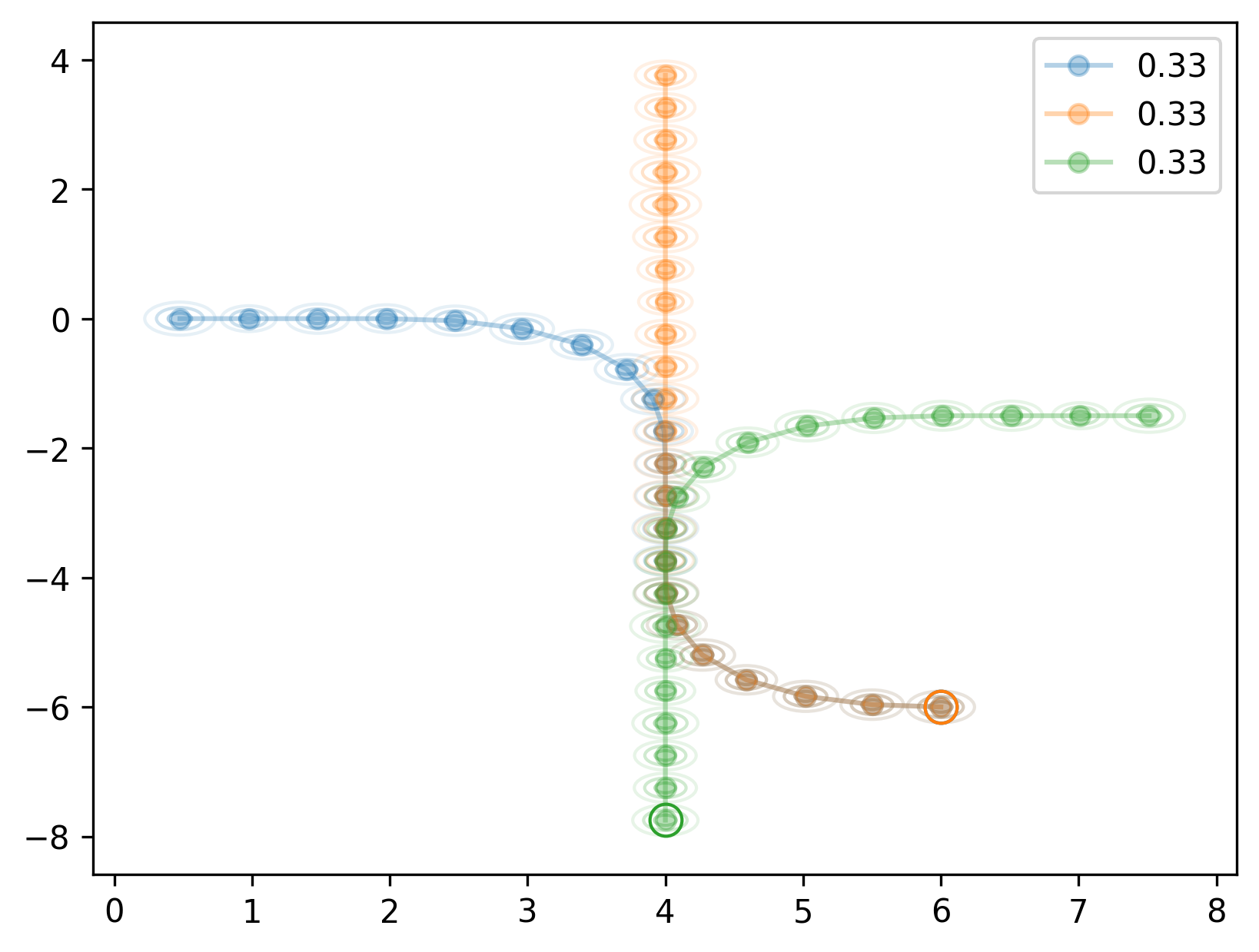}
        \includegraphics[width=0.48\columnwidth]{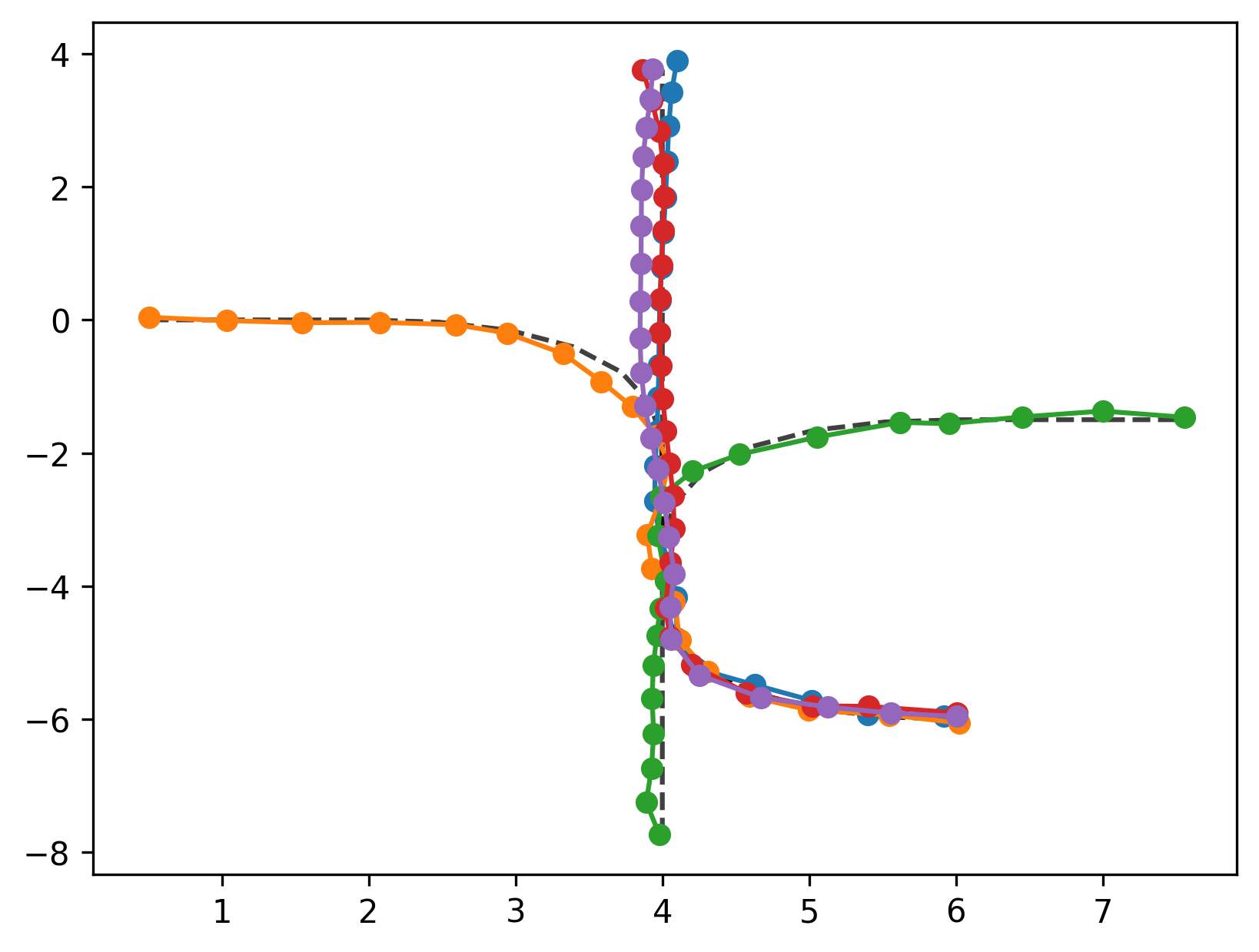}
        \caption{A dataset's prior distribution in terms of a Gaussian mixture covering full trajectories (left) and samples drawn from the prior (right).}
        \label{fig:dataset}
\end{figure}

For this evaluation, a dataset with $3$ paths consisting of straight and curved segments is defined.
All mixture components model trajectories of the same constant movement speed.
The dataset's prior distribution and trajectory samples drawn from the dataset are depicted in Fig. \ref{fig:dataset}.

\paragraph{Training}
As a reference probabilistic trajectory prediction model, an extension to RED \cite{becker2018red}, which enables multi-modal predictions, is used.
The model is trained on a set of $200$ trajectories sampled from the dataset, each capped at a length of $N=19$\footnote{This is the shortest length among the mixture components}, using an NLL-based loss function.
Here, the observation length is set to $N_{obs} = 4$ and the prediction length is set to $N_{pred} = 6$.

\paragraph{Evaluation}
For model evaluation, $20$ additional trajectories are sampled.
In order to evaluate the trained model's performance, the following steps are performed for each test trajectory $\mathcal{X} = \{\mathbf{x}_1,...\mathbf{x}_{19}\}$:
\begin{enumerate}
        \item Extract a sub-trajectory $\mathcal{X}_{N_s} = \{\mathbf{x}_i,...,\mathbf{x}_{i+N_s}\}$ of length $N_s = N_{obs}+N_{pred}$, e.g. $\{\mathbf{x}_2,...\mathbf{x}_{12}\}$. The first $N_{obs}$ points of $\mathcal{X}_{N_s}$ will be denoted as $\mathcal{X}^{obs}_{N_s}$.
        \item Using the dataset's prior mixture distribution $p(X_1,...,X_N)$, for each mixture component find the subset $p_k(X_j,...,X_{j+N_{obs}})$ with the mean sequence $\{\mu^k_j,...,\mu^k_{j+N_{obs}}\}$ closest to $\mathcal{X}^{obs}_{N_s}$.
        %\item For each mixture component then calculate the conditional ground truth distribution $P = p(X_{j+N_{obs}+1},...,X_{j+N_{pred}}|\mathcal{X}^{obs}_{N_s})$ 
        \item Obtain the conditional ground truth distribution\footnote{For the calculation of the conditional weights see e.g. \cite{petersen2008matrix,pml1Book}} $P = \sum_k \pi_{k,\text{pred}|\text{obs}} \cdot p_k(X_{j+N_{obs}+1},...,X_{j+N_{pred}}|\mathcal{X}^{obs}_{N_s})$. 
        \item Pass $\mathcal{X}^{obs}_{N_s}$ through RED in order to obtain a prediction $Q$ for the conditional distribution $P$.
        \item Approximate the distance between $P$ and $Q$ on a per point basis using the sliced Wasserstein distance. Take the average as the final score.
        \item When using the NLL as an additional performance metric, use $Q$ and $\mathcal{X}^{pred}_{N_s}$ for calculation.
\end{enumerate}

\begin{figure}[h]
        \includegraphics[width=0.48\columnwidth]{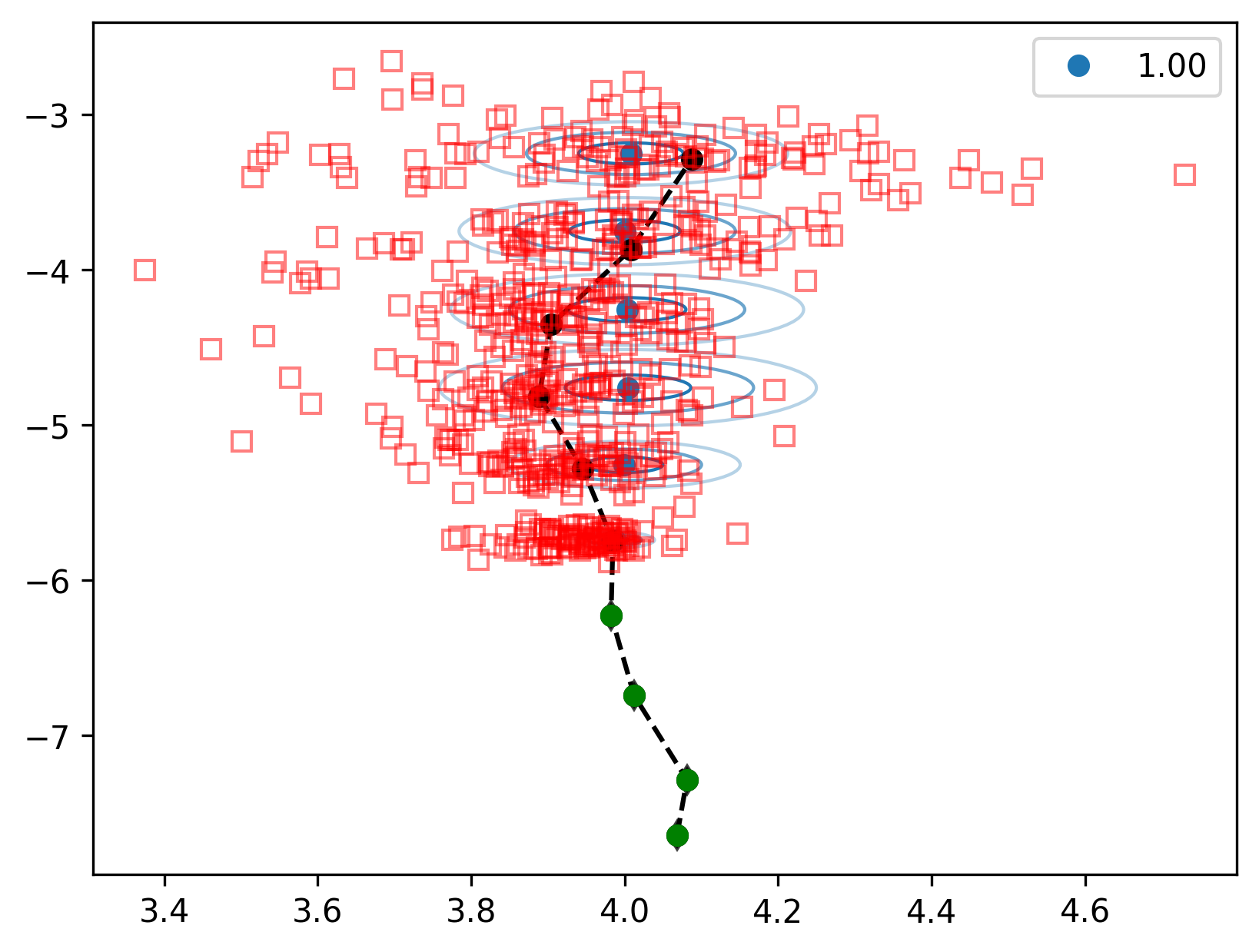}
        \includegraphics[width=0.48\columnwidth]{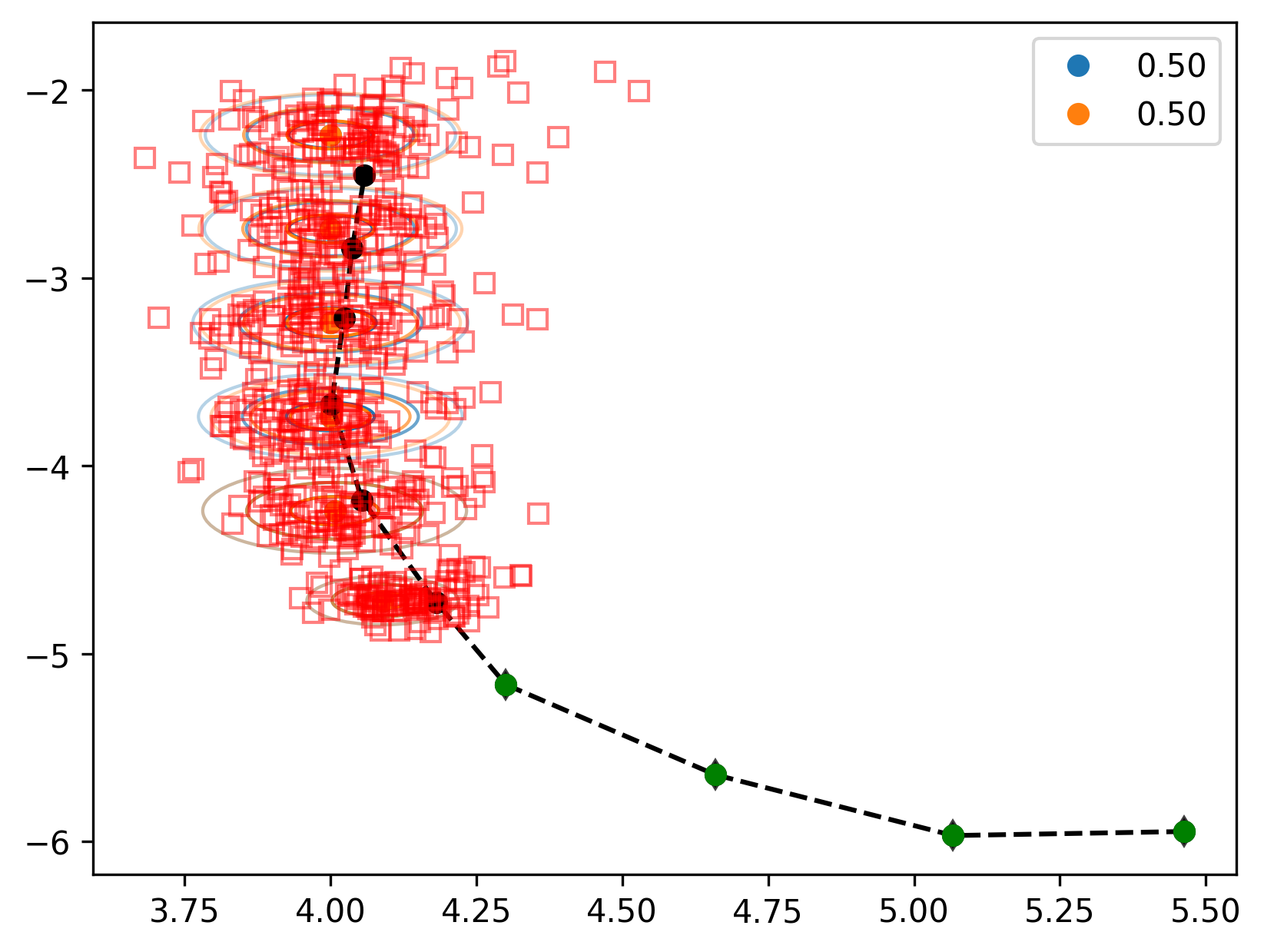}
        \caption{Sample-based predictions generated by the RED predictor for different inputs giving the best score in terms of the NLL (left) and the Wasserstein distance (right). The conditional ground truth is indicated by per point Gaussian mean locations and covariance ellipses obtained through marginalization.}
        \label{fig:nll_wasserstein}
\end{figure}

\paragraph{Merits of using the Wasserstein distance}
Using the above approach and considering all test trajectories, the RED predictor reached an overall score of $-0.74$ when considering the NLL and a score of $0.47$ when considering the Wasserstein distance.
Looking at these scores, one of the main advantages of the Wasserstein distance becomes apparent: \emph{interpretability}.
While the NLL score only allows relative comparisons due to being unbound, a perfect prediction will have a Wasserstein distance of $0$, meaning the score's face value directly allows drawing conclusions about the prediction quality.  
Apart from that, it can be observed that both metrics yield proportional results overall with deviations when ordering model predictions by score. These deviations occur due to the Wasserstein distance being more accurate in scoring the similarity between distributions, especially when considering the distribution's variance.
To give an example, this can be observed looking at the input sample producing the best score in each respective metric depicted in Fig. \ref{fig:nll_wasserstein}, where the Wasserstein distance does a much better job at incorporating variance estimation errors in the output of the prediction model, leading to a prediction closely matching the actual variance receiving a better score than the one with over-estimated variance as is the case for the NLL.
As a final remark, it should be noted that calculating the Wasserstein distance considerably increases the computation time of the evaluation. 
For the test dataset used in this exemplary evaluation, calculating the Wasserstein distances took about $125$ times longer than calculating the NLL scores.  

\section{Summary}
In this paper, a novel approach for generating synthetic trajectory datasets in terms of probability distributions over full trajectories has been proposed.
The approach allows the calculation of arbitrary conditional probability distributions required for a more nuanced evaluation of probabilistic trajectory prediction models by allowing the use of the more expressive Wasserstein distance instead of the negative log-likelihood.
An exemplary evaluation based on this data generation approach has been conducted, which was concluded with a brief discussion on the merits of the Wasserstein distance compared to the negative log-likelihood.

\addtolength{\textheight}{-12cm}   % This command serves to balance the column lengths

%%%%%%%%%%%%%%%%%%%%%%%%%%%%%%%%%%%%%%%%%%%%%%%%%%%%%%%%%%%%%%%%%%%%%%%%%%%%%%%%

\bibliographystyle{IEEEtran}  
\bibliography{bibliography} 

\end{document}